\documentclass[letterpaper, 10 pt, conference]{ieeeconf}  

\usepackage{cite}
\usepackage{amsmath,amssymb,amsfonts}
\usepackage{algorithmic}
\usepackage{graphicx}
\usepackage{textcomp}
\usepackage{xcolor}
\usepackage{subcaption}
\usepackage{changepage}

\IEEEoverridecommandlockouts                              

\title{\LARGE \bf
A Rapid Adapting and Continual Learning Spiking Neural Network Path Planning Algorithm for Mobile Robots
}

\author{Harrison Espino$^{1}$ and Robert Bain$^{2}$ and Jeffrey L. Krichmar, \textit{Senior Member IEEE}$^{1,2}$
\thanks{This work was supported by the Air Force Office of Scientific Research (AFOSR) Contract No. FA9550-19-1- 0306, and by the National Science Foundation (IIS-RI award 1813785 and NSF-FO award IIS-2024633).}%
\thanks{$^{1}$Espino is with the Department of Computer Science, University of California, Irvine, Irvine, CA, USA {\tt\small espinoh@uci.edu}}%
\thanks{$^{2}$Bain is with the Department of Cognitive Sciences, University of California, Irvine, Irvine, CA, USA {\tt\small rkbain@uci.edu}}%
\thanks{$^{1,2}$Krichmar is with the Department of Cognitive Sciences and Department of Computer Science, University of California, Irvine, Irvine, CA, USA {\tt\small jkrichma@uci.edu}}
}

\begin{document}

\maketitle
\thispagestyle{empty}
\pagestyle{empty}

\begin{abstract}

Mapping traversal costs in an environment and planning paths based on this map are important for autonomous navigation. We present a neurobotic navigation system that utilizes a Spiking Neural Network Wavefront Planner and E-prop learning to concurrently map and plan paths in a large and complex environment. We incorporate a novel method for mapping which, when combined with the Spiking Wavefront Planner, allows for adaptive planning by selectively considering any combination of costs. The system is tested on a mobile robot platform in an outdoor environment with obstacles and varying terrain.  Results indicate that the system is capable of discerning features in the environment using three measures of cost, (1) energy expenditure by the wheels, (2) time spent in the presence of obstacles, and (3) terrain slope. In just twelve hours of online training, E-prop learns and incorporates traversal costs into the path planning maps by updating the delays in the Spiking Wavefront Planner. On simulated paths, the Spiking Wavefront Planner plans significantly shorter and lower cost paths than A* and RRT*. The spiking wavefront planner is compatible with neuromorphic hardware and could be used for applications requiring low size, weight, and power.

\end{abstract}

\section{INTRODUCTION}

Finding one’s way around in an ever-changing world is an important part of everyday life. Similarly, robots and other autonomous systems require this capability. Researchers and industry have made considerable progress developing navigation systems. However critical open issues have been identified such as: 1) Generating maps using data from multiple sensors, 2) Continual learning without offline retraining, and 3) Flexibility in the face of a changing environment and different navigational objectives, such as conserving battery usage or avoiding foot traffic \cite{survey3,survey1,survey2}. In the proposed work, we address these issues by developing a novel approach that takes different traversal costs into account when navigating.

We present a spiking neural network navigation system that simultaneously constructs environmental cost maps and uses those maps to plan efficient paths. The system is tested on a ground robot in rugged, varied outdoor terrains with cost maps for obstacles, slope, and for the robot’s effort based on the motor’s current draw. We show that the robot rapidly learns to plan paths that avoid impassable trees and benches with an obstacle cost map, and plans smoother or flatter paths with the current or slope cost map.

Similar works are \cite{hwu2018}, which introduce the Spiking Wavefront Planner and compare it to A* in simplified pre-mapped environments. \cite{krichmar2021} demonstrates E-prop can be used to update the Spiking Wavefront Planner connection delays with experience on simulated environments of a ground robot. The present study marks the first instance in which E-prop and the Spiking Wavefront Planner are combined on a physical robot platform to learn and navigate an environment online and continuously.


The main contributions of this work are as follows:

\begin{enumerate}
    \item We show that our navigation system can simultaneously map complex environments in real time and plan paths over multiple measures of cost. This map is continuously learned online through experience. The robot uses this map to plan trajectories depending on what costs are considered.

    \item In trials with a ground robot, we show that the robot can learn a cost map in a few hundred training steps and several hours of runtime. We also show that the robot can adapt to changes in the environment in just a few trials without taking the system offline.

    \item By exhaustively simulating all paths through our learned costmap, we find that the Spiking Wavefront Planner is the best candidate for minimizing cost and path length between itself, A*, RRT*, and a shortest Euclidean distance planner.
\end{enumerate}

\begin{figure*}[t!]
    \centering
    \includegraphics[width=2.0\columnwidth]{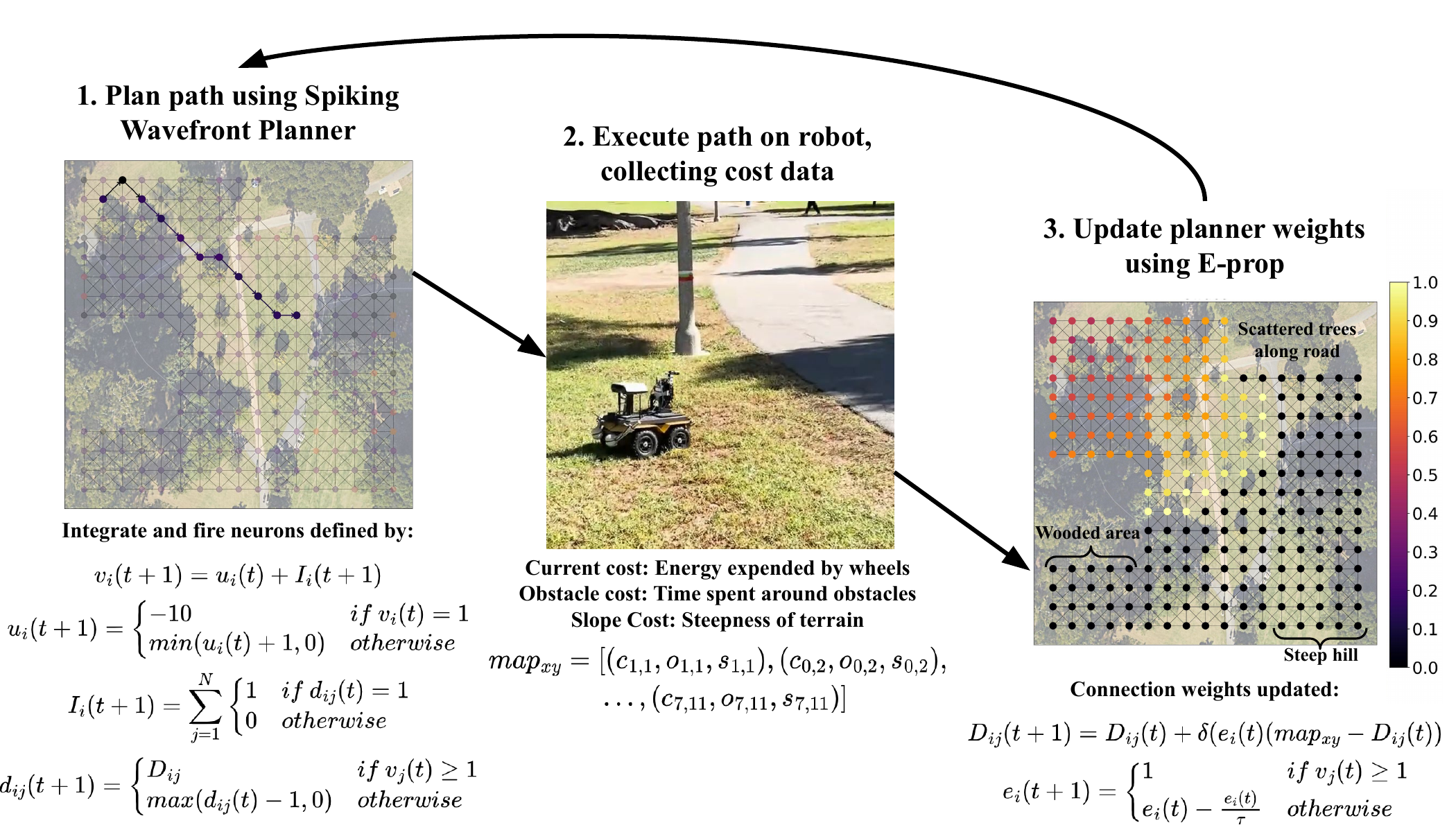}
    \caption{Overview of our navigation system. \textbf{1.} Neurons in the spiking wavefront planner represent locations in space. To plan a path, a spike is induced at the robot's start location and propagates until it reaches the goal location. \textbf{2.} The robot navigates to the goal using the planned path. Internal measurements and environmental sensors track costs when traversing between neuron waypoints. \textbf{3.} Using the learned costs, the planner is updated using E-prop. The figure shows each neuron's eligibility trace, which determines the update magnitude based on how recently the neuron spiked.}
    \label{fig:overview}
\end{figure*}

Figure \ref{fig:overview} provides an overview of the path planning system. The equations in the figure are described in Sections \ref{sec:spikewave} and \ref{sec:eprop}.

\section{RELATED WORKS}

\subsection{Path Planning}

Traditional solutions to path planning often approach the problem by traversing through a graph of costs, or through direct sampling of the state space. Here, we briefly review popular algorithms from both.

A* traverses a graph by considering nodes according to a priority queue sorted by an immediate cost value and a heuristic. Given an admissible heuristic, A* has been proven to provide cost optimal paths \cite{astar}. A* has been utilized for robot navigation as well as applications such as puzzle solving and resource allocation. In unknown environments, the D* algorithm (shortened from Dynamic A*) aims to reach a goal while continually re-planning when new information about the environment is discovered \cite{Dstar}. 

Sampling-based algorithms form a viable path by exploring the space of permissible states. Rapidly exploring random tree (RRT) is one such algorithm designed to efficiently construct a space filling tree until a sequence of valid states between a start and goal state is found \cite{RRT}. It is often employed in high-dimensional trajectory planning problems in which quickly generating a feasible path is more important than generating an optimal one \cite{robot_rrt}.

Many variants of RRT have been created to address problems such as optimality or environmental changes. For the former, RRT* is a variant of RRT which aims to optimize for some cost using a tree rewiring step \cite{rrtstar}. RRTX further builds on this concept for dynamic environments \cite{rrtx}.

\subsection{SLAM}

Mapping of an environment is often accomplished through the process of simultaneous localization and mapping (SLAM). This is a thoroughly explored problem in robotics in which the environmental features and the robot's position in the environment are unknown, and both must be estimated through sensory and self-motion data. Classical solutions to this problem such as extended Kalman filter SLAM (EKF-SLAM) iteratively estimate a posterior probability distribution for the robot pose and landmark positions \cite{EKF}. Other methods use the data as constraints to a graphical network representing the posterior \cite{GRAPHSLAM}, or rely on tracking changes in visual input \cite{ORBSLAM}. In a more biologically-inspired approach, RAT-SLAM integrates visual place recognition and robot position to represent the robot and environment through a grid of "pose cells" \cite{ratslam}.

In these cases, the map consists of geometric environmental features or salient visual features that serve the purpose of aiding localization. Not present are the aspects of steepness, unevenness, or other measures of traversal difficulty meant to aid in navigation. Methods such as \cite{costmapIRL} or \cite{costmapBEVNET} use deep learning approaches such as inverse reinforcement learning or semantic segmentation to generate a costmap of such features for the surrounding environment. Semantic segmentation uses offline training with a dataset of semantically labeled images. Inverse reinforcement learning requires human demonstration with which to infer a reward function.

Solutions to SLAM also assume that the robot's trajectory is decided externally through manual controls or a separate path planning policy. The emerging field of active SLAM combines path planning with SLAM, with the objective of choosing a trajectory which accurately and quickly maps an environment \cite{ASLAM}. The goal of path planning in this paradigm is to formulate a path which minimizes the uncertainty of the SLAM algorithm. However, for some applications, it may be necessary to dynamically consider other goals depending on the context. For instance, an autonomous robot may need to manage battery usage at certain times by planning paths which minimize power consumption. Thus, a navigation system which affords such versatility is needed.

\subsection{Deep Learning Methods}

Deep learning solutions to trajectory planning include imitation learning and reinforcement learning for end to end navigation. Imitation learning methods such as \cite{imit1, imit2, imit3} use human annotated data to mimic expert demonstrations. Due to their reliance on pre-collected data, they do not allow for continuous or autonomous data. Reinforcement learning (RL) methods closely mimic humans learning from interaction with their environments, and have found great success navigating through complex environments. However, they require extensive offline training and expensive onboard computation to run in real-time \cite{BADGR, Manderson, Zhang2018, viking}. As such, there is a need for an online and continuously learning sample-efficient navigation system.

\section{BACKGROUND}

\subsection{Spiking Wavefront Planner}
\label{sec:spikewave}
Here we briefly describe the Spiking Wavefront Planner model. For more details, see \cite{hwu2018, krichmar2021}. 

The spiking wavefront propagation algorithm assumes a grid representation of space, where connections between units represent the ability to travel from one grid location to a neighboring location. Each unit in the grid is represented by simplified integrate and fire neurons. Rather than weights between neurons, the connections between neurons represent a propagation delay, such that a spike signal takes $D$ time steps before being received by a downstream neuron. The activity of neuron $i$ at time $t+1$ is represented by (\ref{eqn:membrane}):
\begin{equation}
    v_{i}(t+1)=u_{i}(t)+I_{i}(t+1),
    \label{eqn:membrane}
\end{equation}
in which $u_{i}(t)$ is the recovery variable and $I_{i}(t)$ is the input current at time $t$. 

The recovery variable $u_{i}(t)$ is described by:
\begin{equation}
    u_{i}(t) =
    \begin{cases}
      \beta & \textit{if} \ v_{i}(t)=1 \\
      \min(u_{i}(t-1)+1,0) & \textit{otherwise}
    \end{cases},
    \label{eqn:recovery}
\end{equation}
such that immediately after a membrane potential spike, the recovery variable starts as a negative value $\beta$ and linearly increases toward a baseline value of 0. For our experiments, $\beta$ is set to -10. We found this to be sufficiently large enough to prevent a spike from reactivating previously visited nodes.

The input current $I$ at time $t+1$ is given by:
\begin{equation}
    I_{i}(t+1)=\sum _{j=1}^{N}
    \begin{cases}
      1 & if \ d_{ij}(t)=1 \\
      0 & \textit{otherwise}
    \end{cases},
    \label{eqn:synin}
\end{equation}
such that $d_{ij}(t)$ postpones the integration of input, $I$, from neighboring neuron $j$ to neuron $i$. This delay is given by:
\begin{equation}
    d_{ij}(t+1) =
    \begin{cases}
      D_{ij} & \textit{if} \ v_{j}(t)\ge1 \\
      \max(d_{ij}(t)-1,0)        & \textit{otherwise}
    \end{cases}.
    \label{eqn:delay}
\end{equation}

The value of $D_{ij}(t)$ is the propagation delay between neurons $i$ and $j$, and denotes the expected cost of traveling from location $i$ to $j$. This is initialized to 1 for all values. Cost is an open parameter, which could depend on a number of variables. In the present paper, multiple measures of cost are measured simultaneously, which is explained in more detail in section \ref{cost}.

\subsection{E-Prop}
\label{sec:eprop}
The E-Prop learning rule was developed to learn sequences in recurrent spiking neural networks by using an eligibility trace to implement backpropagation through time to minimize a loss function \cite{Bellec2020}. The present work used E-Prop to learn a map of the environment, which is represented by a recurrent spiking neural network, based on the sensed cost of traversal. For path planning purposes, the active neurons after a wave propagation are eligible for updates. An eligibility trace based on time elapsed since the wave reaches the goal destination dictates the eligibility. E-Prop is applied to the delay $D_{ij}$ between neuron $i$ and $j$ along the traversed path. 

\begin{equation}
    D_{ij}(t+1)=D_{ij}(t)+\delta(e_{i}(t)(m_{xy}-D_{ij}(t)),
    \label{eqn:update}
\end{equation}
where $\delta$ is the learning rate, set to 0.5, $e_{i}(t)$ is the eligibility trace for neuron $i$, and $m_{xy}$ represents the cost observed from the robot's sensors at location $(x,y)$, which corresponds to neuron $i$. This rule is applied for each of the neighboring neurons, $j$, of neuron $i$. The loss in Eqn. \ref{eqn:update} is $m_{xy}$ - $D_{ij}$.

The eligibility trace for neuron $i$ is given by (\ref{eqn:eligibility}):
\begin{equation}
    e_{i}(t+1) =
    \begin{cases}
      1 & if \ v_{j}(t)\ge1 \\
      e_{i}(t)-\frac{e_{i}(t)}{\tau}        & otherwise
    \end{cases},
    \label{eqn:eligibility}
\end{equation}
where $\tau$ is the rate of decay for the eligibility trace, set to 25. An example of the eligibility trace from a planned path can be seen in the right panel of Fig \ref{fig:overview}.

To determine a path from the robot's current location to a destination, a signal is sent originating from the neuron corresponding to the robot's current location. This signal propagates based on the delays $D$ to the origin neuron's neighbors. This is repeated for each of these neurons and their neighbors until a signal reaches the neuron corresponding to the destination. The origin of this signal is recursively traced backwards until the neuron representing the current location is reached. This sequence of neurons represents the path of least cost given the robot's information about the environment.

\section{Experimental Setup}

\subsection{Robot Platform and Environment}

\begin{figure}[t!]
    \centering
    \includegraphics[width=0.9\columnwidth]{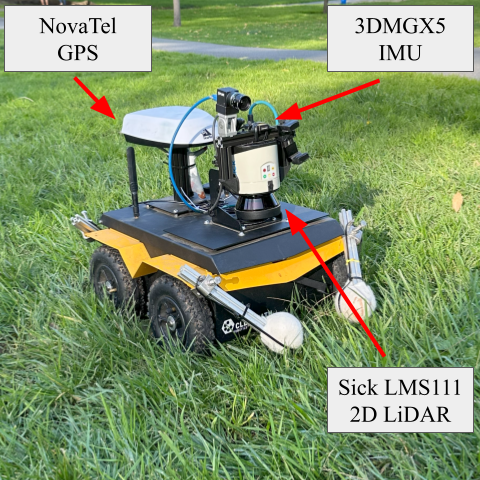}
    \caption{The Clearpath Jackal robot.}
    \label{fig:jackal}
\end{figure}

The robot platform used in the experiments was the Jackal Unmanned Ground Vehicle from Clearpath Robotics (Fig. \ref{fig:jackal}). The Jackal is capable of navigating through difficult, uneven terrain. To localize the robot in its environment and determine the distance to waypoints we used a NovaTel GPS unit. To determine the robot's heading and bearing, we used the Lord Microstrain 3DMGX5 inertial measurement unit (IMU).

\begin{figure}[t!]
    \centering
    \includegraphics[width=0.9\columnwidth]{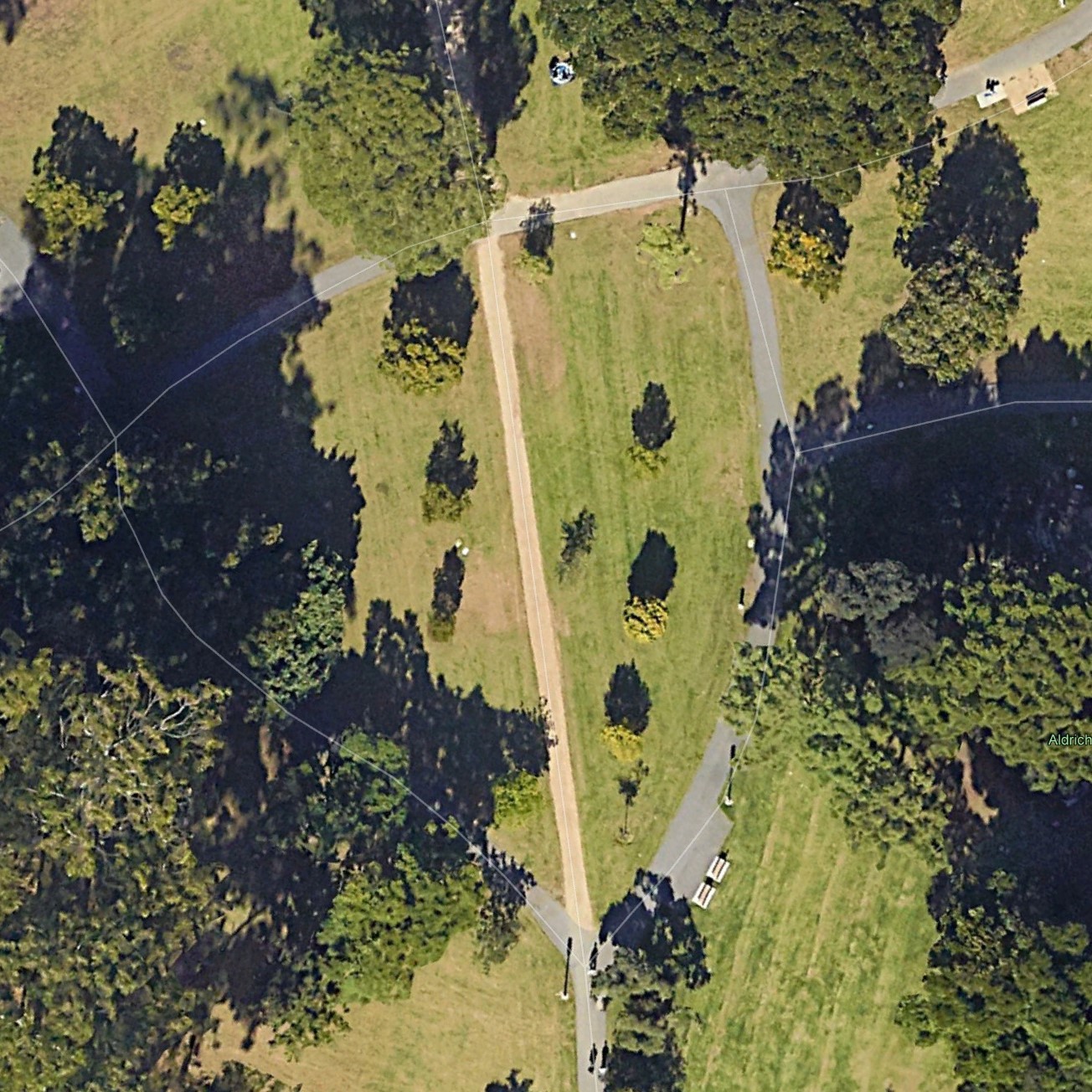}
    \caption{Top down view of the Aldrich Park environment. Imagery \copyright Google}
    \label{fig:sat_nowps}
\end{figure}

We tested our navigation system in Aldrich park: A hilly park located at the center of of the University of California, Irvine. A top down view of the environment can be seen in Fig. \ref{fig:sat_nowps} A 17x17 grid, 5.1 meters apart were used as waypoints (see Fig. \ref{fig:costmap}). This distance was chosen based on the precision of the GPS unit of 1.2 meters. Terrain in the environment was hilly and varied between thick grass, paved road, and dirt road. The environment contained a number of trees which served as obstacles for the robot in addition to foot traffic along the road. The robot explored this environment for 350 trials over 4 days and approximately 12 hours of total runtime. Some sections of the grid were removed from consideration, as they were completely intraversable due to large root structures that could not be detected by the LiDAR. 

\subsection{Evaluation and Comparisons}

Using the learned costmap, we compared the Spiking Wavefront Planner to the  RRT* \cite{rrtstar} and A* \cite{astar} path planning algorithms. For RRT*, the extension of the tree was constrained to be only in the direction of waypoints. This was necessary, as the costmap only contains delays for movement in the cardinal and ordinal directions. For A*, we used the common heuristic of shortest Euclidean distance.

We tested paths generated by both of these algorithms as well as the Spiking Wavefront Planner on 25 randomly selected paths whose start and end waypoints were at least 3 waypoints apart to ensure meaningful choices in path planning. During traversal, we collected measures corresponding to each of the learned costs.

To evaluate the effectiveness of the Spiking Wavefront Planner in navigating our cost maps, we also exhaustively simulated all potential start and end locations with a minimum separation of 3 waypoints. This was similarly done for A*, RRT* and a naive planner which generated a path with the shortest euclidean distance.

\section{METHODS}

\subsection{Cost Measures}
\label{cost}

We considered three measures of cost: 
\begin{enumerate}
    \item \textbf{Current Cost:} The amount of current from the battery to the left and right wheels. This was determined internally using the status messages automatically published by the Jackal. For each incoming current reading, we considered the minimum value between the left and right side. This was to prevent spikes in current caused by turning from influencing the cost, which may happen due to trajectory recalculation and not from features in the environment.

    \item \textbf{Obstacle Cost:} The presence of obstacles during traversal. To recognize obstacles we used the SICK LMS111 2D LiDAR, which has an aperture angle of 270 degrees and an angular resolution of 0.5 degrees. Objects were considered obstacles if the LiDAR detected a number of consecutive data points below a threshold of 2 meters. The cost before normalization was calculated as the fraction of time spent with obstacles in the field of view en route to the waypoint. The Jackal avoided an obstacle by turning away from it until the obstacle was out of view. In the case that a collision was still eminent (such as if part of the obstacle was too high or low for the LiDAR), manual controls were used to direct the Jackal past the obstacle. This was a rare occurrence, and was only necessary due to benches in the environment lying above the field of view.

    \item \textbf{Slope Cost:} The slope of the ground  was based on the pitch and roll readings from the IMU. Because flat ground is measured as 0 radians of rotation about both axes, this cost (before normalization) was calculated as the sum of the rotation about the pitch and roll axes.
\end{enumerate}

Additionally, a cost representing completely intraversable locations was included, which incurred cost only when the robot was unable to reach a waypoint in the allotted time. In this case, a maximum cost of 10 was assigned to the delays into this waypoint to discourage its use in future paths.

In order to convert costs to values suitable for training the Spiking Wavefront Planner, sensor data needed to be converted into integers representing learnable delay values. We chose 10 as the maximum delay value to maintain a fast network response during wave propagation. To obtain normalization constants for each measure, the Jackal was driven between waypoints prior to training. Minimum and maximum values were calculated as two standard deviations below and above the mean. Outliers during training were clamped prior to normalization.

Each of these costs are maintained individually for a given neuron's delay values. To combine costs into a single map, the delay values for each chosen cost are added together. This value is then normalized between 1 and 10 again across all neurons to maintain a fast network response.

\subsection{Environment Mapping}

A single trial with the robot proceeded as follows. First, an end point was randomly determined using a Levy Flight distribution. This distribution is commonly used to model foraging patterns in animals \cite{yang2010}. The Levy Flight distribution will tend to focus search in a local area while occasionally jumping to a distant area. This was a better exploration strategy than a more random search pattern, such as Brownian motion.

Next, the Spiking Wavefront Planner planned a path from the robot's current location to the end point by setting the activity $v_i$ of the starting neuron to 1 and simulating the neuron behavior as in equations (\ref{eqn:membrane}) through (\ref{eqn:delay}). The delay values $d_{ij}$ for each neuron used a costmap combining all measures of cost. The eligibility trace generated by the Spiking Wavefront Planner was used to update the delays with E-prop as in equation (\ref{eqn:update}). The robot then navigated between waypoints determined by the generated path.

To reach a waypoint, the robot oriented to the direction of the waypoint by rotating until the robot's heading (orientation with respect to a global reference frame) matched the desired bearing (orientation with respect to the waypoint). When the heading was suitably close, the robot would proceed towards the waypoint. If at any point during traversal the difference between the bearing and the heading exceeded $\frac{\pi}{12}$, the robot would stop to rotate to the correct orientation before proceeding. We found this value to be a suitable level of precision for our IMU.

The trial was complete once all waypoints were reached, or the robot was unable to reach a waypoint after 45 seconds. In the latter case, the robot would return to the previous waypoint. After each trial, the delays of the costmap were saved. The saved delays were used in simulated experiments to analyze possible paths taken by the robot under different path planning algorithms.

\section{Results}

\subsection{Environment Mapping}

\begin{figure}[t!]
    \centering
    \includegraphics[width=1.0\columnwidth]{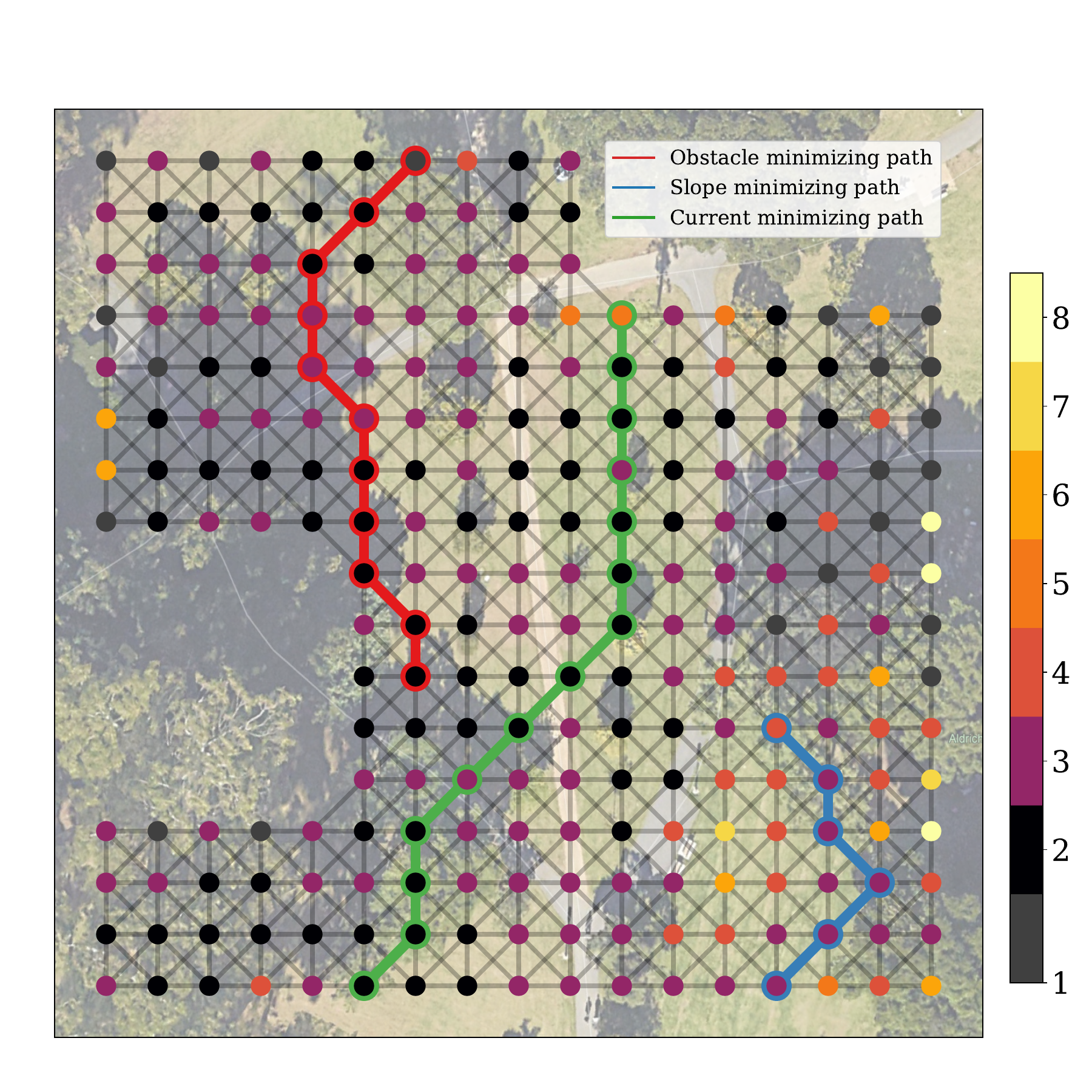}
    \caption{The costmap for all costs added and normalized after learning. Nodes are colored according to the mean of the delays $D$ from other nodes. Example paths minimizing current drawn, obstacles encountered, and steepness are colored green, red, and blue, respectively.}
    \label{fig:costmap}
\end{figure}

The costs determined from the robot's exploration of Aldrich park are shown in Fig. \ref{fig:costmap}. Nodes are colored according to the mean of delays $D$ from other nodes using all costs combined. Candidate paths minimizing current drawn, obstacles encountered, and steepness are colored green, red, and blue, respectively.

Using the current cost criteria, we found similar values between the grass, dirt road, and pavement, indicating they were similarly traversable in terms of energy consumption. Areas with slightly elevated current cost included waypoints around trees and at intersections between terrain. We speculate this is because the transition from grass to pavement or to dirt road caused increased unevenness, and consequently, stress on the motors. The current minimizing path took a relatively straight path visiting a minimal amount of waypoints (green line in Fig. \ref{fig:costmap}).

The obstacle cost criteria produced sparser costs, with higher costs at tree and bench locations. Foot traffic along the sidewalk also resulted in higher obstacle costs. Occasionally, due to the unevenness from the road or from sloped areas, the robot was also at an angle steep enough to briefly detect the ground as an obstacle. In the obstacle minimizing path (red line in Fig. \ref{fig:costmap}), the robot avoided a row of trees by taking a slightly longer route.

Because of a steep hill, there was a higher slope cost in the bottom right quadrant of the cost map. When planning a path through this hill, the robot traversed up the hill, moved along a flat ridge for most of its route, and then down the hill to reach the goal (blue line in Fig. \ref{fig:costmap}).

\begin{figure}[t!]
    \centering
    \includegraphics[width=0.9\columnwidth]{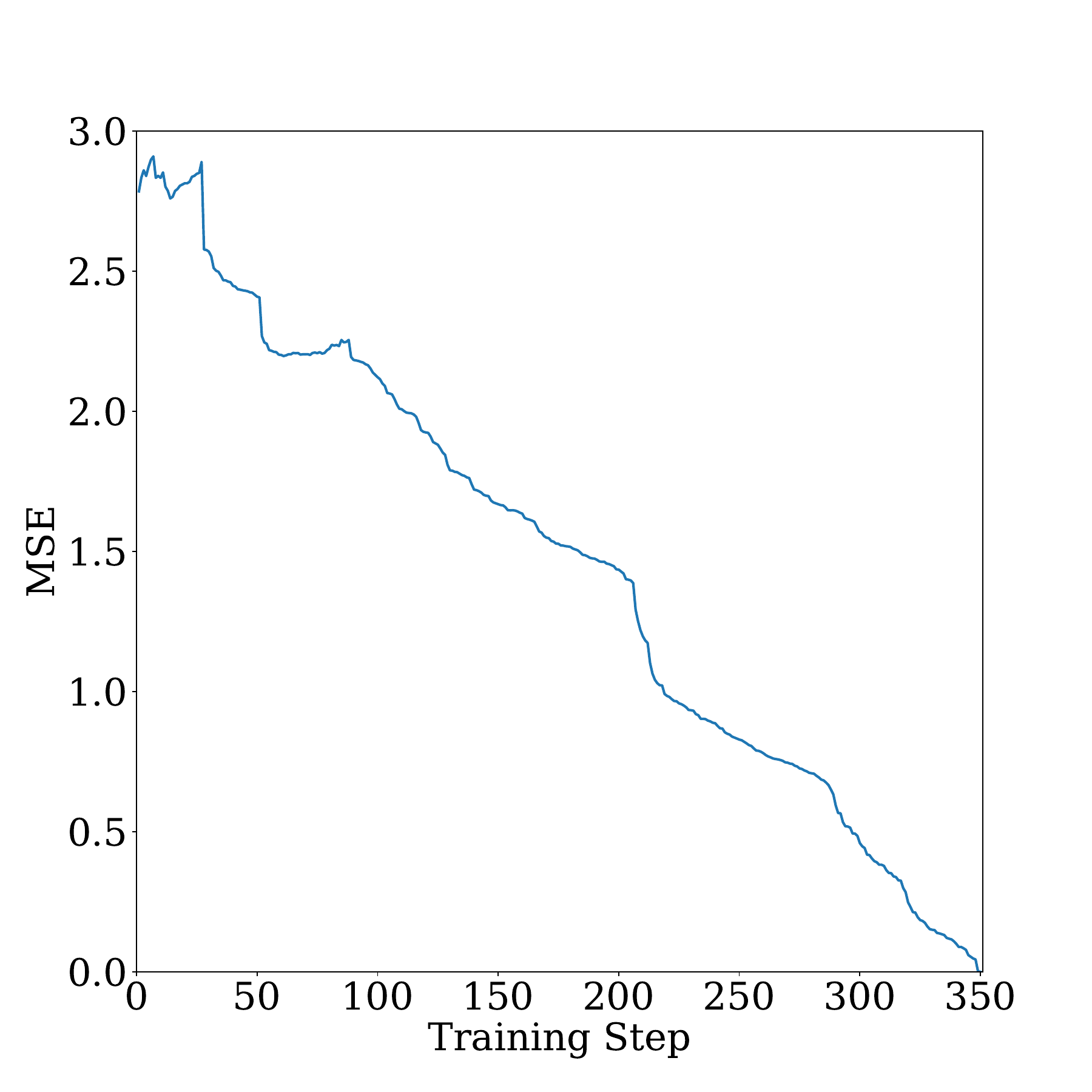}
    \caption{Mean squared error between the delays $D$ of the model at each training step and the final learned costs of the model.}
    \label{fig:mse}
\end{figure}


\begin{figure}[t!]
    \centering
    \includegraphics[width=1.0\columnwidth]{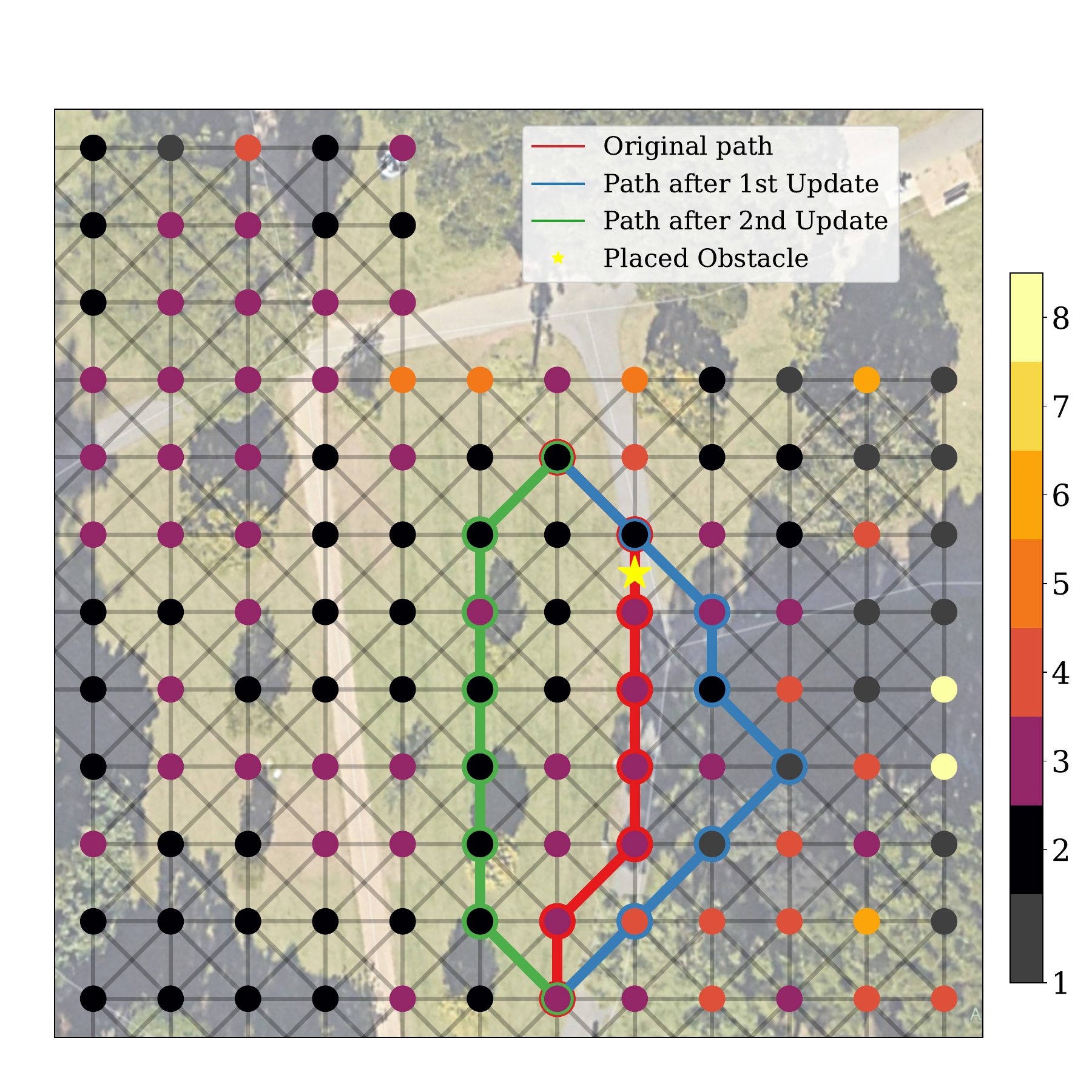}
    \includegraphics[width=1.0\columnwidth]{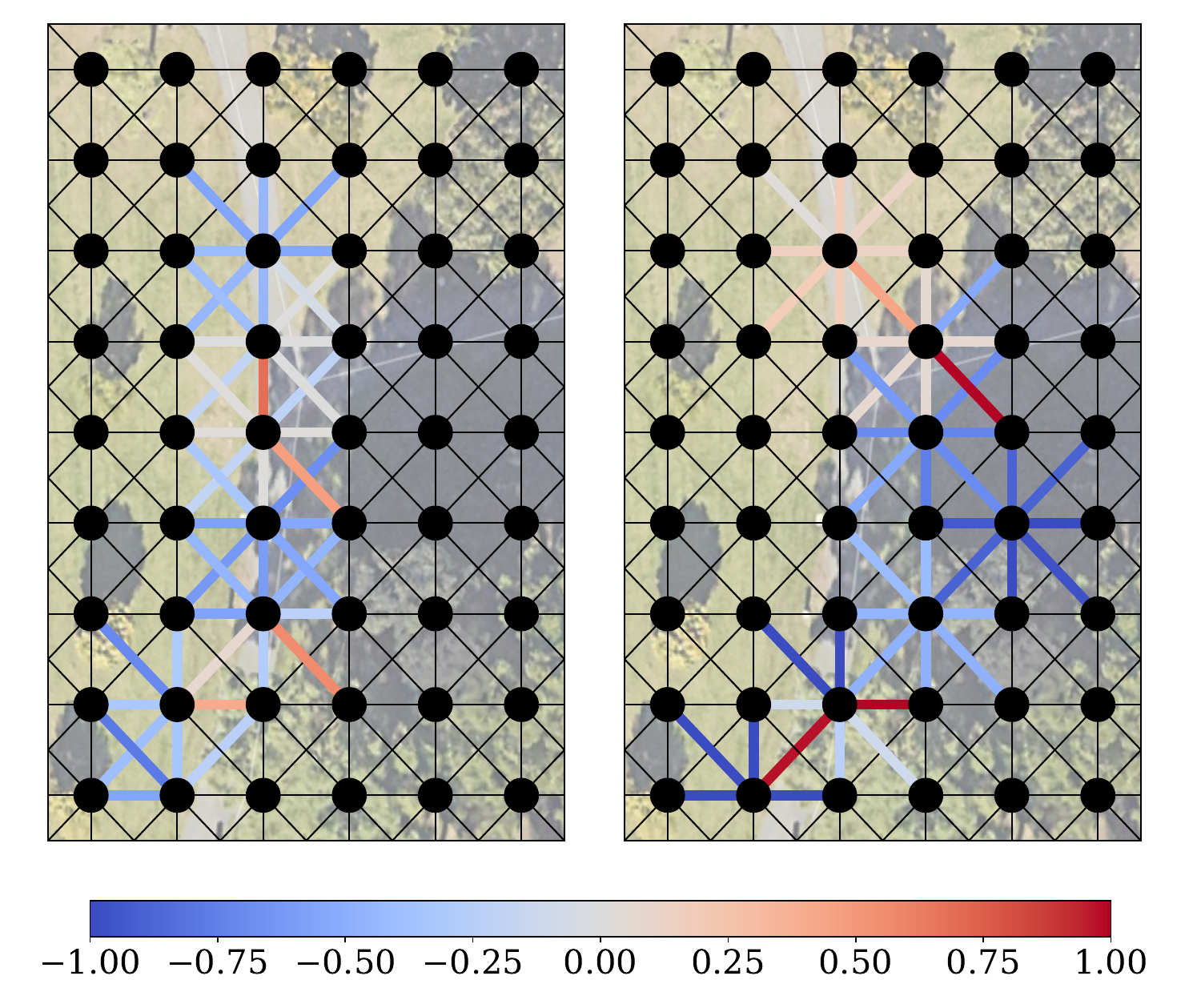}
    \caption{Top image shows example paths demonstrating adaptation after multiple experiences in a changing environment. Red, green, and blue lines outline the planned path after zero, one, and two updates to the model, respectively. The location of manually placed obstacles are marked with a yellow star. Bottom images shows changes to the model delays after the first update (left) and second update (right). Colored edges indicate the extent of the delay change between the two neurons.}
    \label{fig:adaptive}
\end{figure}

\subsection{Continual Learning and Adaptation}

Continual learning is realized through E-prop's ability to generate an increasingly accurate map of the environmental costs as the robot explores its environment. Fig. \ref{fig:mse} compares the delays of the Spiking Wavefront Planner after each executed path with that of the final learned model. Mean squared error steadily declined, which suggests that each training step is improving upon the last. 

Continual learning facilitated rapid adaptation in the face of environmental changes. To demonstrate adaptation, we placed an obstacle in the robot's path. After one or two experiences in this new situation, the robot updated its cost map with this information. Fig. \ref{fig:adaptive} shows the new paths taken by the robot after the first and second experience with this obstacle.

The initial path (shown in red) travels along waypoints on the road, however the presence of the obstacle incurs high obstacle cost, as evidenced by high weight changes at the obstacle location. After a single update, the second path (shown in blue) avoids the obstacle but travels through a sloped hill area to do so, incurring a higher slope cost. This can be seen from the high weight changes when traversing up or down the hill. With one final update, the final path (shown in green) then travels away from the road along flat grass to reach the goal, avoiding both the placed obstacle and the hill.

\subsection{Comparisons with Existing Path Planning Algorithms}

\begin{table}[t]
    \begin{minipage}{\columnwidth}
        \caption{Comparison between RRT*, A*, and Spiking Wavefront Planner (SWP) on paths taken by robot.}
        \centering
        \begin{tabular}{l|ccc}
            \textbf{Planner}       & \textbf{RRT*}                                            & \textbf{A*}                                              & \textbf{SWP (ours)} \\ \hline
            \textbf{Path Length}   & \begin{tabular}[c]{@{}c@{}}60.23*\\ \end{tabular}  & \begin{tabular}[c]{@{}c@{}}46.98\\ \end{tabular}    & 46.6         \\ \hline
            \textbf{Current drawn}  & \begin{tabular}[c]{@{}c@{}}468.74\\ \end{tabular} & \begin{tabular}[c]{@{}c@{}}382.82\\ \end{tabular} & 415.89       \\ \hline
            \textbf{Obstacles encountered} & \begin{tabular}[c]{@{}c@{}}4.45*\\ \end{tabular}   & \begin{tabular}[c]{@{}c@{}}3.95\\ \end{tabular}   & 2.06         \\ \hline
            \textbf{Slope}    & \begin{tabular}[c]{@{}c@{}}7.52\\ \end{tabular}     & \begin{tabular}[c]{@{}c@{}}7.72\\ \end{tabular}     & 7.57       \\
            \hline
            \textbf{Normalized Cost}    & \begin{tabular}[c]{@{}c@{}}24.63\\ \end{tabular}     & \begin{tabular}[c]{@{}c@{}}21.83\\ \end{tabular}     & 21.67 
        \end{tabular}%
        \captionsetup{justification=raggedright,singlelinecheck=false,margin={1.5cm,0cm}}
        \caption*{\footnotesize \textit{* denotes p\textless0.05; t-test with Bonferonni correction}}
        \label{tab:realpaths}
    \end{minipage}
\end{table}

In tests with the physical robot in paths in the environment (Table \ref{tab:realpaths}), the Spiking Wavefront Planner planned significantly shorter paths and minimized obstacle costs better than RRT*. Both planners performed similarly on other costs. A* was most comparable to the Spiking Wavefront Planner, as neither path length nor measures of cost were significantly different between the two algorithms. The lack of significance in many of these metrics could be due to not enough sample routes.

\begin{table}[t]
    \begin{minipage}{\columnwidth}
        \caption{Comparison between RRT*, A*, a Naive planner, and Spiking Wavefront Planner (SWP) on simulated paths.}
        \centering
        \begin{tabular}{l|cccc}
        \textbf{Planner}       & \textbf{RRT*}                                            & \textbf{A*}                                               & \textbf{Naive}                                              & \textbf{SWP (ours)} \\ \hline
        \textbf{Path Length}   & \begin{tabular}[c]{@{}c@{}}59.01*\\ \end{tabular} & \begin{tabular}[c]{@{}c@{}}51.73*\\ \end{tabular} & \begin{tabular}[c]{@{}c@{}}47.73*\\ \end{tabular} & 50.91        \\ \hline
        \textbf{Normalized Cost} & \begin{tabular}[c]{@{}c@{}}22.07*\\ \end{tabular} & \begin{tabular}[c]{@{}c@{}}19.55*\\ \end{tabular} & \begin{tabular}[c]{@{}c@{}}21.85*\\ \end{tabular}    & 19.36       
        \end{tabular}%
        \captionsetup{justification=raggedright,singlelinecheck=false,margin={0.6cm,0cm}}
        \caption*{\footnotesize \textit{* denotes p\textless0.05; t-test with Bonferonni correction}}
        \label{tab:simpaths}
    \end{minipage}
\end{table}

To overcome the small sample size, we simulated all possible paths of length greater than 3 ($n$ = 57086) on the learned costmap itself. The results of this are shown in Tab. \ref{tab:simpaths}. The Spiking Wavefront Planner significantly outperformed RRT* and A* on minimizing cost. Additionally, paths generated by the Spiking Wavefront Planner were significantly shorter than RRT* and A*. The paths were longer than the naive planner, however this was expected as the naive planner travels the shortest euclidean distance regardless of cost.

\begin{figure}[t!]
    \centering
    \includegraphics[width=0.9\columnwidth]{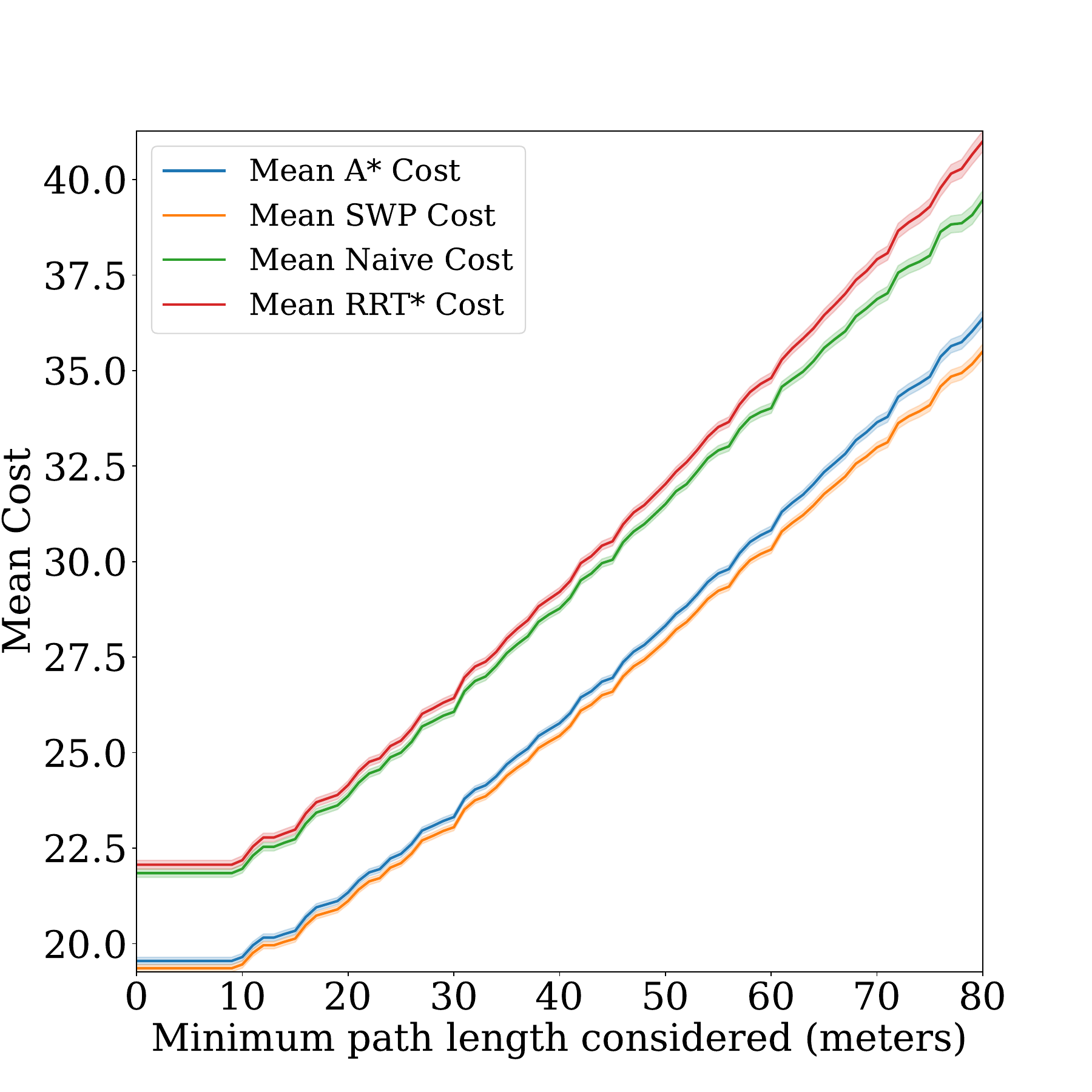}
    \caption{Difference in performance for each path planning algorithm as the length of simulated paths is increased. Y-axis is the mean cost of the paths. X-axis is the minimum path length considered. Shaded areas represent 99\% confidence interval.}
    \label{fig:improvement}
\end{figure}

Fig. \ref{fig:improvement} illustrates how the comparison between algorithms changes over increasingly longer path lengths. The Spiking Wavefront Planner performed the best, and RRT* performed the worst in terms of cost. As the paths got longer, the disparity between algorithm performance became more pronounced.

\section{CONCLUSIONS AND FUTURE WORK}

The Spiking Wavefront Planner with E-Prop learning can rapidly learn traversal costs for navigation and can adapt to change without lengthy offline retraining. It demonstrated shorter and more cost effective paths than other path planning algorithms. Although A* may perform similarly with a tailored heuristic, this requires careful consideration of how to estimate the future measure of cost at a given location. Moreover, a different heuristic may be necessary for each cost or combination of costs. By contrast, the Spiking Wavefront Planner can be universally applied to a costmap regardless of what combination of costs are considered.

Our navigation system is not without its limitations. Currently, costs are obtained through experience only and cannot generalize between waypoints or environments. Depending on the learning rate, it may take multiple passes between the same two waypoints to properly learn an accurate cost. It has also been shown in previous work that the Spiking Wavefront Planner is computationally slower than A* \cite{hwu2018}.

These limitations serve as avenues for future research. Our current implementation does not use vision, however computer vision techniques for self-labeling such as those found in \cite{BADGR, viking} could be used to estimate the cost of current and nearby trajectories during traversal. It has also been shown that incorporating biologically-inspired memory replay can improve exploration speed and adaptation to changes in the environment \cite{Espino}. Lastly, while the Spiking Wavefront Planner was tested on traditional hardware in the present work, its spiking properties mean it could be deployed on power efficient neuromorphic hardware as demonstrated in \cite{fischl2017}. It remains to be seen whether this will also offer improvements in computational efficiency.

In summary, this work demonstrates our efficient navigation system for off-road navigation that learns continuously from interaction with its environment in real-time, without the need for multiple rounds of training and deployment or expensive hardware. The system learns multiple measures of cost in parallel, and can plan paths that minimize such costs or any combination of them when traversing the environment. Through our real world and simulation results, we determined that these paths are shorter and more cost effective than A* and RRT*. The simplicity of the software stack supports much future development, perhaps using context from a camera, memory replay \cite{Espino}, more robust obstacle detection and avoidance, and neuromodulation \cite{JinweiNeuromod} to control combining costmaps.

\bibliographystyle{ieeetr}
\bibliography{egbib}

    
    
    

\addtolength{\textheight}{-12cm}   









\end{document}